\documentclass{article}
\usepackage{arxiv}

\usepackage{natbib}
\usepackage{times}
\usepackage{latexsym}
\usepackage{url}
\usepackage[hidelinks]{hyperref}
\usepackage{booktabs}

\usepackage[T1]{fontenc}

\usepackage[utf8]{inputenc}

\usepackage{graphicx}
\usepackage{tabularx}
\usepackage{xspace}

\usepackage{multirow}
\usepackage{amssymb}
\usepackage{authblk}

\usepackage{caption} 
\usepackage{rotating}

\newcommand{\repo}{\url{https://github.com/dkorenci/oppositional-discourse}}

\newcommand{\rqone}{\textbf{RQ1}\xspace}
\newcommand{\rqtwo}{\textbf{RQ2}\xspace}
\newcommand{\rqthree}{\textbf{RQ3}\xspace}

\title{What Distinguishes Conspiracy from Critical Narratives? \\ A Computational Analysis of Oppositional Discourse}
\date{}

\newcommand{\eqcontribsymbol}{\dag}
\makeatletter
\newcommand{\printfnsymbol}[1]{%
  \textsuperscript{\eqcontribsymbol}%
}
\makeatother

\newbox{\orcid}\sbox{\orcid}{\includegraphics[scale=0.06]{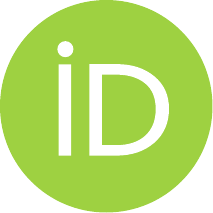}} 
\author[1,2]{%
	\href{https://orcid.org/0000-0003-4645-2937}{\usebox{\orcid}\hspace{1mm}Damir Koren\v{c}i\'c\printfnsymbol{2}\thanks{\texttt{name.surname@irb.hr}}}%
}
\author[3,4]{%
	\href{https://orcid.org/0000-0003-1169-0978}{\usebox{\orcid}\hspace{1mm}Berta Chulvi\printfnsymbol{2}}%
}
\author[5]{%
	\href{https://orcid.org/0009-0003-8827-0215}{\usebox{\orcid}\hspace{1mm}Xavier Bonet Casals}%
}
\author[1]{%
	\href{https://orcid.org/0000-0001-6955-9249}{\usebox{\orcid}\hspace{1mm}Alejandro Toselli}%
}
\author[5]{%
	\href{https://orcid.org/0000-0003-0089-940X}{\usebox{\orcid}\hspace{1mm}Mariona Taulé}%
}
\author[1,6]{%
	\href{https://orcid.org/0000-0002-8922-1242}{\usebox{\orcid}\hspace{1mm}Paolo Rosso}%
}
\affil[1]{Universitat Politècnica de València, Spain}
\affil[2]{Ru{\dj}er Bo\v{s}kovi\'c Institute, Croatia}
\affil[3]{Symanto Research, Spain}
\affil[4]{Universitat de València, Spain}
\affil[5]{CLiC - Universitat de Barcelona, Spain}
\affil[6]{ValgrAI - Valencian Graduate School and Research Network of Artificial Intelligence, Spain}

\renewcommand{\headeright}{}
\renewcommand{\undertitle}{}
\renewcommand{\shorttitle}{A Computational Analysis of Oppositional Discourse}

\title{What Distinguishes Conspiracy from Critical Narratives? A Computational Analysis of Oppositional Discourse}

\begin{document}

\maketitle
\def\thefootnote{\eqcontribsymbol}\footnotetext{Equal contribution} 
\def\thefootnote{\arabic{footnote}}
\def\thefootnote{}\footnotetext{Accepted version of the paper: \url{https://doi.org/10.1111/exsy.13671}} 
\def\thefootnote{\arabic{footnote}}

\begin{abstract}
The current prevalence of conspiracy theories on the internet is a significant issue, tackled by many computational approaches. However, these approaches fail to recognize the relevance of distinguishing between texts which contain a conspiracy theory and texts which are simply critical and oppose mainstream narratives.
Furthermore, little attention is usually paid to the role of inter-group conflict in oppositional narratives. We contribute by proposing a novel topic-agnostic annotation scheme that differentiates between conspiracies and critical texts, and that defines span-level categories of inter-group conflict. We also contribute with the multilingual XAI-DisInfodemics corpus (English and Spanish), which contains a high-quality annotation of Telegram messages related to COVID-19 (5,000 messages per language). We also demonstrate the feasibility of an NLP-based automatization by performing a range of experiments that yield strong baseline solutions. Finally, we perform an analysis which demonstrates that the promotion of intergroup conflict and the presence of violence and anger are key aspects to distinguish between the two types of oppositional narratives, i.e. conspiracy vs. critical.
\end{abstract}


\section{Introduction}
\label{sec:intro}

Conspiracy Theories (CTs) are complex narratives that attempt to explain the ultimate causes of significant events as cover plots orchestrated by secret, powerful, and malicious groups \citep{Douglas_2023}. 
The automatic detection of CTs in written texts has recently gained popularity, and the problem is commonly framed as a binary classification task, which can be detailed more thoroughly with fine-grained approaches corresponding to multi-label or multi-class classifications \citep{giachanou_conspi_2021, moffitt_2021_conspiracy, pogorelov_mediaeval2021, pogorelov_mediaeval2022}. 

However, existing approaches do not distinguish between critical and conspiratorial thinking. This distinction has important implications for automatic content moderation: if models do not differentiate between critical and conspiratorial thinking, there is a high risk of pushing people toward conspiracy communities.
Namely, labeling a text as conspiratorial when it is actually opposing mainstream views, could potentially lead people who were simply asking questions closer to conspiracy communities. As several authors from the field of social sciences suggest, a fully-fledged conspiratorial worldview is the final step in a progressive ``spiritual journey" that sets out by questioning social and political orthodoxies \citep{SUTTON2022,Funkhouser2022,FranKs2017}. Additionally, recent research \citep{Phadke2021} has shown that the level of interaction with conspiratorial users is the most important feature for predicting whether or not users join conspiracy communities.

Another gap in the computational analysis of conspiratorial written texts is the failure to address the role that intergroup conflict (IGC) plays in conspiratorial and critical narratives. Applying the concept of intergroup conflict to the study of  language consists of identifying a particular approach to the framing of events, one that emphasizes the hostility between groups, typically by using the ``\textit{us} versus \textit{them}'' narrative, and by fueling the perceived injustice and threat to the ingroup (for a recent review of the IGC, see \citealp{BOHM2020947}). The relevance of connecting the study of conspiracy theories with the theoretical framework of IGC stems from the increasing and potentially violent involvement of conspiratorial communities in political processes. 
This involvement suggests that the purpose of CTs is to reinforce group dynamics and to coordinate action \citep{WAGNEREGGER2022}. Therefore, the tools that enable an IGC-based analysis of conspiratorial texts could offer valuable insights for both social sciences and content moderation.  

Motivated by the described issues, we propose a novel annotation scheme that distinguishes 
between conspiracy and critical texts, and defines important categories of oppositional narratives (Figures \ref{fig:conspi} and \ref{fig:critical} in subsection \ref{sec:binaryannot} contain examples of this annotation scheme).
In addition to the core elements of conspiracy narratives that have been widely used in the literature, such as \textit{Agents} (the conspirators) and \textit{Victims}, the proposed annotation scheme identifies two additional  categories: \textit{Facilitators} (those collaborating with the agents, such as the media) and \textit{Campaigners} (those that unmask the conspiratorial agenda). These types of actors are considered as key players in IGC: on the one hand, the facilitators are tangible targets with whom real conflict is possible (in contrast to the agents, which tend to operate in secret),
and on the other hand, the campaigners are those who show their opposition to the facilitators and try to persuade the victims to join the cause. 

To summarize, distinguishing between critical discourse on the one hand, and inflammatory, violent, and destabilizing narratives on the other, is important for democratic societies which seek to promote freedom of speech and ensure a civilized exchange of opinions. This distinction is also crucial in order to avoid excessive censorhip.

In order to tackle the described issues, we focused on oppositional Telegram messages related to the COVID-19 pandemic, and constructed the multilingual (English and Spanish) XAI-DisInfodemics corpus, annotated with a novel annotation scheme. We then proceeded to tackle several Natural Language Processing (NLP) tasks related to annotation tasks we have carried out during the construction of the corpus. These include 
a binary classification task of distinguishing between conspiratorial and critical texts, and a multi-label annotation task at the span level to detect the main elements of both kinds of oppositional narratives (Section \ref{sec:dataset}).  
Both the code used in the experiments and the data have been made available.\footnote{\repo}

In this paper, the following research questions are addressed: 
\begin{itemize}
    \item \rqone: Can the critical-vs-conspiracy classification task be solved successfully with state-of-art NLP methods?
    \item \rqtwo: Can narrative elements, including the IGC-related elements, be automatically detected?
    \item \rqthree: Is the proposed annotation scheme a useful analytical tool for interdisciplinary research on computational linguistics and social science? 
\end{itemize}

This article is organized in the following sections: in Section (\ref{sec:rw}) we first give an overview of the current research context. In Section (\ref{sec:dataset}) we proceed to describe the XAI-DisInfodemics corpus, the annotation scheme, and the annotation process. 
In Section \ref{sec:ml-experiments} we tackle \rqone and \rqtwo by analyzing the performance of fine-tuned transformers, and we demonstrate that the \textit{critical} vs. \textit{conspiracy} distinction can be detected with a high level of accuracy. We also demonstrate that the detection of oppositional narrative categories is a more challenging task, especially at the fine-grained span level, and that the binary detection of these categories is an approach that can achieve high accuracy for the majority of the categories. 
In Section \ref{sec:css-experiments} we tackle \rqthree by conducting experiments using variables which take into account the \textit{Campaigners} and \textit{Facilitators} categories from the annotation scheme.
We show that an important distinction between critical and conspiratorial narratives is that the latter promote intergroup conflict more intensely and convey more words related to anger and political violence. Finally, in Section \ref{sec:conclusions} we outline the conclusions and future work. 

Our work makes four main contributions: (1) an annotation scheme based on a novel approach to oppositional text analysis that focuses on both the \textit{conspiracy} vs. \textit{critical} distinction, and on the IGC-related narrative elements; (2) the multilingual XAI-DisInfodemics corpus, consisting of Telegram messages related to COVID-19; (3) several machine learning experiments that examine the tractability of the scheme-derived tasks and yield strong baseline solutions for future applications; (4) the analyses that demonstrate the usefulness of the proposed approach in computational social sciences. 

\section{Related Work}
\label{sec:rw}

The detection of conspiracy theories in written text has been approached by several authors using NLP techniques both in Computational Linguistics and in Computational Social Sciences (CSS). \citet{bessi_personality_2016} used a text scaling method to map conspiratorial text to personality traits and to analyze conspiracies. Topic modeling has been used to extract and analyze common themes in conspiracy texts \citep{klein_conspitopic_2018, samory_government_2018}. \citet{samory_government_2018} use syntactic parsing in order to extract ``motifs'' (agent-action-target triplets) and analyze patterns of their occurrence.
\citet{holur_which_2022} performed detection of \textit{insider} and \textit{outsider} entities in conspiracy texts by automatic labeling of noun phrases. \citet{giachanou_conspi_2021} used psychological and linguistic features to classify and analyze social media users that spread conspiracies. In contrast, \citet{levy_investigating_2021} analyzed the capacity of large language models to generate conspiracies. In recent years, there has been a lot of work on NLP approaches to detect and analyze COVID-19 conspiracies in written texts. \citet{moffitt_2021_conspiracy} developed a classifier of conspiracy tweets and used it for propagation analysis. Two recent MediaEval challenges which focused on classification of conspiracy texts \citep{pogorelov_mediaeval2021, pogorelov_mediaeval2022} led to a number of approaches demonstrating that the state-of-the-art architecture is a multi-task classifier \citep{peskine2021, peskine_2022, korencic_mediaeval_2023} based on CT-BERT \citep{muller_ctbert_2020}. 

While other authors also focused on methods that analyze conspiratorial text at the span level \citep{holur_which_2022, samory_government_2018, introne_mapping_2020}, the main contribution of our work is that we take into account the importance of intergroup conflict.

\citet{holur_which_2022} recognize the relevance of IGC: ``conspiracy theories and their constituent threat narratives share a signature semantic structure: an implicitly accepted Insider group; a diverse group of threatening Outsiders''. However, they do not define the \textit{insider} and \textit{outsider} labels in the sense of conflicting social groups. The label \textit{insider} is defined in the sense of \textit{familiar} or \textit{pleasant}, and the label \textit{outsider} as the opposite. These binary classes are in contrast with the more fine-grained and IGC-oriented categories that we define in Section \ref{sec:oppositional-annot}. Additionally, \citet{holur_which_2022} restrict the labeling to noun phrases, while we allow the spans to range from a single word to whole clauses.   

\citet{samory_government_2018} propose an unsupervised method to extract agent-action-target triplets describing conspiracy narratives, but it fails to capture the IGC-related aspect of the texts. In addition, the subject-predicate-object triplets can fail to match with more abstract agent-action-target triplets. In contrast, we use a flexible span definition that allows for different types of actors and other narrative elements. 

\citet{introne_mapping_2020} propose a span-level scheme of six categories (event, actor, goal, action, consequence, target), and use it to analyze 236 messages from anti-vaccination forums. They distinguish between conspiracy theories and conspiratorial thinking, a category that implies only passive support. 
This distinction is not based on annotations grounded in theory, but on the requirement of all the categories being present in a given text. However, in practice fewer elements can convey a conspiracy theory in a very strong manner. 

Several corpora have been built to study conspiracy theories or phenomena related to them.
\citet{memon2020misinfo} constructed a dataset of 4,700 tweets with 17 misinformation-related categories, 
including one for conspiracy spreading tweets. 
\cite{song2021classification} collected a corpus of 1,293 debunked COVID-19 claims and annotated them with a number of misinformation categories, including \textit{conspiracy theory}. 
\cite{holur_which_2022} developed the CT5K corpus, consisting of 5,000 COVID-19 conspiracy messages, collected from several social platforms, which they used for the previously described experiments with insider-outsider classification.

\citet{uscinski_conspiracy_2011} collected a dataset consisting of letters sent to a mainstream US publication, and labeled them as either containing a conspiracy or not. The large scale LOCO corpus \citep{miani_loco_2021} contains 96,743 texts from a diverse collection of mainstream and conspiracy media outlets. The texts are enriched with website metadata and auto-generated topics. COCO is a corpus of 3,495 texts containing COVID-19 conspiracies \citep{langguth_coco_2023}. The texts were manually annotated with a fine-grained classification scheme encompassing conspiracy sub-topics.

The recent literature review by \citet{mahl_conspiracy_2022} shows that the interest about conspiracy theories in online environments has been exponentially rising within the Social Sciences, and that 80\% of work on this topic was done on written content, with about a third of the articles using methods for automated content analysis. This demonstrates the relevance of our research, as we provide a new corpus, conceptual tools, and NLP methods that enable more sophisticated analyses than traditional methods of textual content analysis, which rely on techniques such as text scaling and word counting.

\section{Dataset: XAI-DisInfodemic Corpus}
\label{sec:dataset}

This section introduces the XAI-DisInfodemic corpus, a multilingual (English and Spanish) corpus consisting of 5,000 annotated Telegram messages each. These messages contain oppositional non-mainstream views on the COVID-19 pandemic, classified into two categories: critical messages and conspiratorial messages. The aim of this corpus is to distinguish these two types of oppositional narratives, which are not equivalent and have very different effects on society: while conspiracy theories about COVID-19 call into question the whole system of social organisation to fight against a disease, critical thinking may express disagreements that are tractable within the system of social organisation of a given community. 
These messages have been annotated in two consecutive phases, which correspond to two separate annotation tasks. 
Firstly, we present the construction of the corpus (subsection 3.1), which includes the source and selection criteria of the messages. Secondly, we describe the annotation methodology for the first task (subsection 3.2) and for the second task (subsection 3.3). For each task, we specify the tagset, the annotation criteria, the annotation process, and the IAA agreement tests. Finally, we present the process of anonymization of the corpus (subsection 3.4).

\subsection{Corpus Construction}
The messages which constitute the corpus were obtained from public Telegram channels in which users tend to post messages which oppose the mainstream discourse related to the COVID-19 pandemic. Even though the social network X (formerly known as Twitter) is commonly used as a source for this kind of research \citep{langguth_coco_2023}, we decided to opt for Telegram messages for several reasons: firstly, because they tend to be longer than X messages, which allows for a deeper analysis of oppositional narratives. Secondly, Telegram has much less restrictive moderation policies, so it is easier for conspiracy theorists to post their opinions more freely \citep{vergani2022hate}.

\begin{figure}[htp]
    \centering
    \includegraphics[width=4cm]{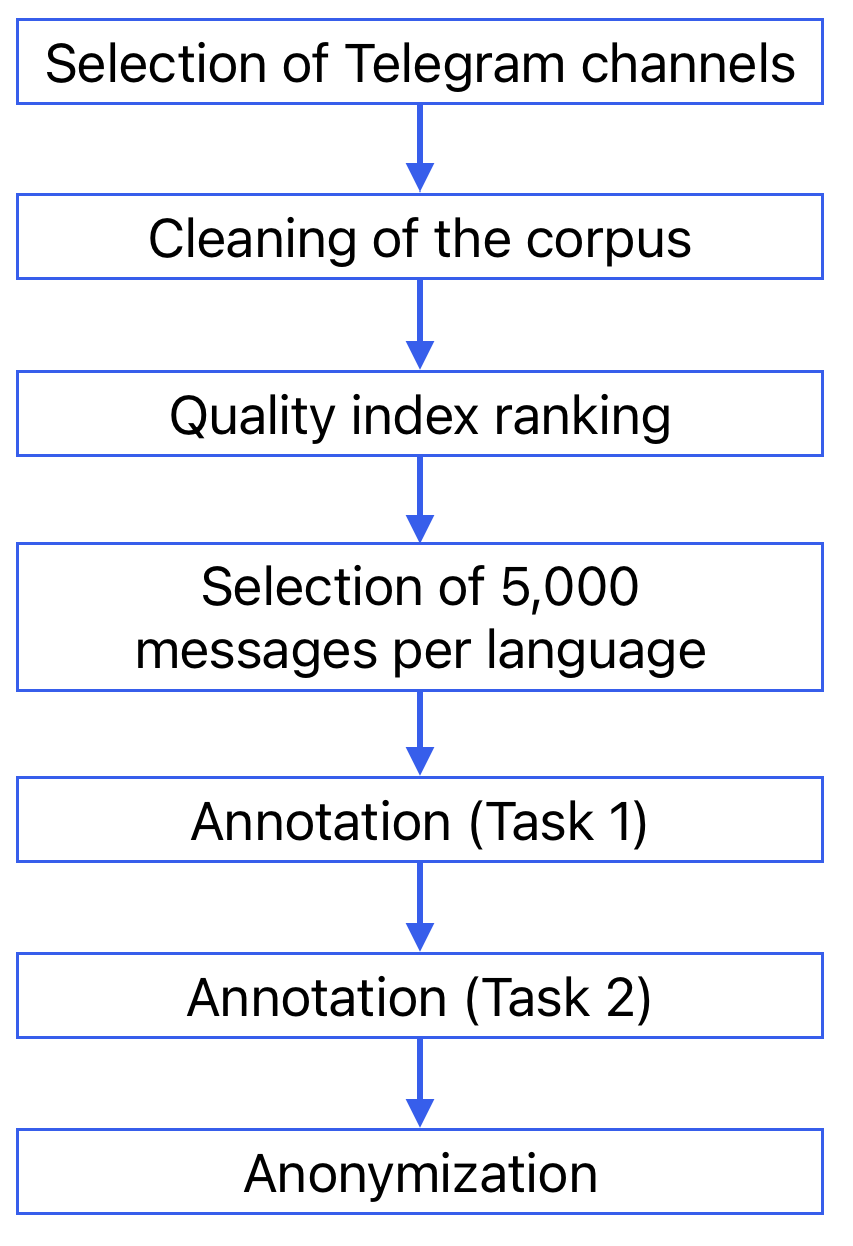}
    \caption{The process of corpus construction.}
    \label{fig:corpusprocess}
\end{figure}

The collection of the messages was carried out as depicted in Figure \ref{fig:corpusprocess}. Firstly, we created a list of 2,273 public Telegram channels, either in English or in Spanish, that contain oppositional non-mainstream views on the COVID-19 pandemic. Locating such channels is challenging since they are familiar only to members of closed groups aligned with channels' attitudes. The first step in the creation of this list was to use a ‘snowballing’ procedure \citep{Chaim2008}
: one of the researchers became a member of several social networks related to conspiracy theories. Specifically, he joined the extreme right-wing network GAP, several Facebook groups, and started to follow on X (former Twitter) a number of users who were openly critical of the government's health measures, and/or public disseminators of conspiracy theories. After that, the sequence of characters ``t.me/'' was used to locate additional oppositional Telegram channels in the aforementioned social networks, since Telegram channels are often identified with the short URL ``t.me/'', followed by the name of the channel. The final step was to repeat this last procedure within the Telegram channels themselves, until no new channels were found.

Then the messages were downloaded from the channels and filtered by removing duplicates, short texts (i.e. those containing fewer than 12 tokens), and texts with a large proportion of URLs and mentions. Finally, each message was ranked using a quality index. This quality index is composed of criteria capturing the activity of the channel and the message length.
Concretely, the quality score of a message was the sum of the scores of the following sixcriteria: (1) the audience of the channel, (2) the number of authors in the channel, (3) the mean of messages by author, (4) the standard deviation of the mean of messages by author, (5) the number of messages in the channel, and (6) the number of words in the message.
Once the messages were sorted using the quality score they were manually reviewed, starting from the top, to discard the messages not related to COVID-19. This way 5,000 comments related to COVID-19 and having highest quality scores were selected for each language. The details of the process of filtering and ranking of the messages can be found in the supplementary material.
Table \ref{corpstats} contains basic statistics of the final English and Spanish corpora. 

 \begin{table*}[ht] 
 \centering
   \begin{tabular}{lccccccc}
     \hline
     Language & Avg. & Std. dev & Min. & Q1 & median & Q3 & Max. \\
     \hline
     Spanish & 128 & 123 & 23 & 49 & 98 & 148 & 766 \\
     English & 265 & 528 & 12 & 32 & 65 & 266 & 4,108 \\     
    \hline
 \end{tabular}
 \caption{Statistics of the text length, measured in number of words (whitespace separated tokens), for English and Spanish corpora: the average, the standard deviation, the minimum, the first quartile, the median, the third quartile, and the maximum.}
 \label{corpstats}
 \end{table*}

\subsection{Conspiracy vs. Critical Annotation}
\label{sec:binaryannot}
The aim of the first annotation task was to identify whether a message hints at the existence of a conspiracy theory, or whether it is criticizing mainstream views on COVID-19 but do not suggest the existence of a conspiracy. To that purpose, we developed an annotation scheme with two binary labels, \textit{Conspiracy} and \textit{Critical}. Each of these labels can have either the value 1 if the message contains the corresponding feature, or the value 0 if it does not contain it. These labels are mutually exclusive (i.e. no message can have the value 1 in both labels simultaneously). Messages annotated with the value 0 in both labels were not included in the final corpus (but are available if required). The total amount of these discarded messages are 767 in English and 360 in Spanish. This low amount of ``neutral'' messages is due to the fact that the corpus was extracted from Telegram channels which tend to contain oppositional messages rather than general domain ones.

For the first annotation task, we developed an annotation scheme to differentiate between texts hinting at the existence of a conspiracy, and those criticizing mainstream views on COVID-19 but without suggesting the existence of a conspiracy. 
A comment was labeled as \textit{Conspiracy} if any of these five criteria were met: 

\begin{enumerate}
\item The message frames COVID-19 or a related public health strategy as the result of the agency of a small and malevolent agent (either known or unknown) operating in the dark, with the intention to carry out a hidden plan. This plan could be to control society, to physically or mentally manipulate citizens or to impose a false narrative, among others \citep{Douglas_2023}. For example: "\textit{Whether you believe it to be psychological, physical or both, vaccines have been weaponized as a tool for population control. This much is now obvious}".  It can also be considered a conspiracy if the vaccine is described as some sort of intentional `poison', and the official vaccination campaign is portrayed as a deliberate murder or as an experiment carried out by an (explicit or implicit) agent, who is purposely trying to harm society. For example: ``\textit{Pfizer knew the Covid jabs were killing 3\% of those who received it. This video is an excellent and succinct summary of the dangers of the “poison death jab” and the musical chairs game big pharma media and govt agents have been playing on us.} [...]''. The hidden plan may be presented in the form of a `join the dots' narrative, i.e. unrelated facts are presented as being connected with the intention of `proving' the existence of a powerful agent who is deliberately responsible of large-scale events. For example: ``\textit{Looks like \#Covid and all the evil caused by the vaccines were just a test run for the larger plan of locking us all into areas codes... Denial is no longer an option}''.

    \item The message implies that the pandemic is not real, and that "a false version of reality has been promoted or protected by the conspirators and their unwitting stooges" \citep{Douglas_2023}. For example: ``\textit{They have already started their bull shit - so the vaccines didn't work - the last lock downs didn't work - but these globalist scum want it all to happen again as their immune compromised sheep have no immune system and are filling the hospital beds and dying} [...]''.    

    \item 
    The message suggests that those who publicly mock or deny the existence of a conspiracy theory and its negative consequences are actually involved in the secret plan and are collaborating with the agents.
    For example: ``[...] \textit{Las vacunas experimentales COVID-19 están causando mucho daño. En este caso un infarto un día después del pinchazo. Y los sanitarios lo saben.} [...]''. (English translation: \textit{The experimental COVID-19 vaccines are causing a lot of damage. In this case, a heart attach just one day after the jab. And health staff know about it}).

    \item A distinction is made between the ingroup (i.e. those who know the truth and reveal the conspiracy, `us') and the outgroup (i.e. those who remain ignorant, society at large, `them'). The outgroup is homogenized and its members are perceived as being manipulated and/or ignorant. For example: "\textit{\#NeverForget all of this fake staging they did - to be used as demagoguery for the pedantic sheep to use against one another, shaming their neighbours into mass compliance for a ‘global pandemic’ hoax}". However, the simple presence of a narrative `them' is not necessarily an indication of a conspiracy theory: when two opposing ideas are presented, it is quite usual to associate each idea to its own champion social entity, but there is a difference between establishing an opposition with another social group (in the sense of `their behavior is irrational' or `they refuse to listen', which is a feature of critical thinking) and presenting `them' as an enemy (which is a feature of conspiracy theories). 

\end{enumerate}

On the other hand, messages with critical thinking also contain elements of an oppositional narrative, but focus on the way the pandemics has been managed by authorities, as well as the effects and/or efficacy of the vaccine. A message was labeled as \textit{Critical} if any of these four criteria were met: 
\begin{enumerate}
    \item The official vaccination strategy and/or the efficiency of vaccines is called into question (e.g. due to the fact that they were developed and commercialized in a very short period of time). This critical stance does not rely on esoteric explanations and the comment does not deny the validity of scientific protocols, but it rather provides what the author considers to be factual data or objective evidence (e.g. the author’s first-hand experience, second-hand accounts from acquaintances or other online users, etc.). For example: "\textit{YouTube has blocked renowned physician Dr. Drew Pinsky from going live on its platform after he showed a vaccine-related eye injury he incurred after receiving an experimental Covid jab.} [...]". 

    \item The comment establishes a link between the vaccine and a number of side effects which are not included in the mainstream narrative. This causal relationship is sometimes referred to as a `crime' due to a (perceived) unusually high mortality after the vaccination campaigns, but the comment does not imply any sort of deliberate intentionality. For example: "\textit{The Spike Protein’s Impact: “Vaccination may cause our overall immune system to fail to fight against such bad things” - Shigetoshi Sano, M.D., PhD., Professor and Chair, Kochi University School of Medicine}".
   
    \item Asking for information about potential links between the vaccine and its effects on public health  is also considered to be a feature of critical thinking, due to the context in which such requests are taking place (i.e. it is assumed that by making use of this kind of channels the author expects a different answer from what could be obtained at an official health center). For example: "\textit{Alguien del grupo tiene o puede tener contacto con médicos/sanitarios que vivan en Reino Unido (zona de Escocia) y estén ayudando/aconsejando a vacunados? Necesitaría algún teléfono, grupo de Telegram... lo que sea que sirva de ayuda. Gracias}." (English translation: "\textit{Does anyone in the group have or may have contact with doctors/healthcare providers who live in the UK (Scotland area) and are helping/advising vaccinated? I would need any phone numbers, Telegram groups... anything that would be helpful. Thanks}"). However, if the comment is simply asking for or giving information of any other kind (e.g. a phone number to contact someone, where a given product can be bought, etc.), it is not considered to include either a conspiracy theory or critical thinking, and is therefore annotated with the value 0 on both labels. 
  
    \item The comment supports a deliberate rejection of the vaccine and/or expresses dissatisfaction with the negative consequences that this rejection entails (e.g. social discrimination, mobbing, administrative constraints, etc.). Oftentimes, the group of the vaccinated and the group of the unvaccinated are presented as antagonistic, but there is no allusion to an agent with a hidden agenda. For example: "\textit{Si vives en España,la vacunación es voluntaria. Si los superiores les dicen que se inoculen, que te lo dé por escrito, como no lo van hacer, grábalo y te servirá ante un juicio. Hay muchas personas como tú. RESISTE}" (English translation: "\textit{If you live in Spain, the vaccine is optional. If your superiors tell you to get jabbed, ask them to give it to you in writing, since they are not going to, record it and it will help you in court. There are many people like you. RESIST}").
\end{enumerate}

A key aspect that serves to distinguish conspiratorial thinking from critical thinking is that the latter emphasises impersonal, systemic causes while the former focuses on the agency of a secret group. Recurrent themes in critical thinking are the questioning of the efficacy of vaccines, criticism of the pharmaceutical industry, criticism of the mandatory nature of vaccines, and the side effects of COVID-19 vaccines. 

The annotation process of this first task was carried out by two teams of annotators (one for the Spanish corpus and one for the English corpus), all of them expert linguists. Each team was constituted of 3 different annotators, either native  or highly proficient speakers in the corresponding language (therefore, each message was annotated by 3 different people). The annotators carried out the task independently, and on a weekly basis meetings were held to discuss disagreements among the annotators and also with the help of two senior annotators (the social psychologist who created the taxonomy, and an expert in corpus linguistics). This process lasted about 5 months.
In order to assess the quality of the annotation, we performed a number of tests. For the English corpus, the IAA in terms of Krippendorf's $\alpha$ is of 0.79 for \textit{Conspiracy} messages and 0.60 for \textit{Critical} messages, and the average observed percentage of agreement between the three annotators is of 91.4\% for \textit{Conspiracy} and 80.3\% for \textit{Critical}. For the Spanish corpus, Krippendorf's $\alpha$ is of 0.80 for \textit{Conspiracy} and 0.70 for \textit{Critical}, and the average observed percentage of agreement is of 90.9\% for \textit{Conspiracy} and 84.9\% for \textit{Critical}. Finally, the Gold Standard was obtained by means of a majority vote (i.e. agreement between at least two annotators).

\subsection{Oppositional Narrative Annotation}
\label{sec:oppositional-annot}

The aim of the second annotation task was to identify the main elements of intergroup conflict (IGC) which appear in oppositional narratives. In order to do that, we selected the span of text in which each element appears and assigned it the corresponding label. The tagset includes six labels, which can be applied both to messages containing a conspiracy theory and messages containing critical thinking. In this section, we describe the labels which compose the tagset, and also present the criteria for span delimitation.

\begin{figure*}[h!]
    \centering
    \begin{tabular}{c}
        \textbf{Conspiracy Theory} \\
        \includegraphics[width=\linewidth]{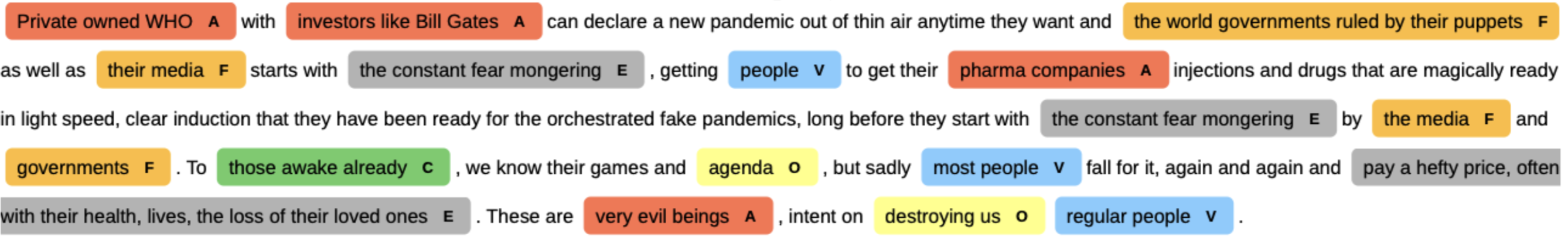} \\
    \end{tabular}
    \caption{A \textit{Conspiracy} message annotated with elements of oppositional narrative: \textit{Agents} (A), \textit{Facilitators} (F), \textit{Campaigners} (C), \textit{Victims} (V), \textit{Objectives} (O), \textit{Negative Effects} (E).}    
    \label{fig:conspi}
\end{figure*}

\begin{figure*}[h!]
    \centering
    \begin{tabular}{c}
        \textbf{Critical Thinking} \\
        \includegraphics[width=\linewidth]{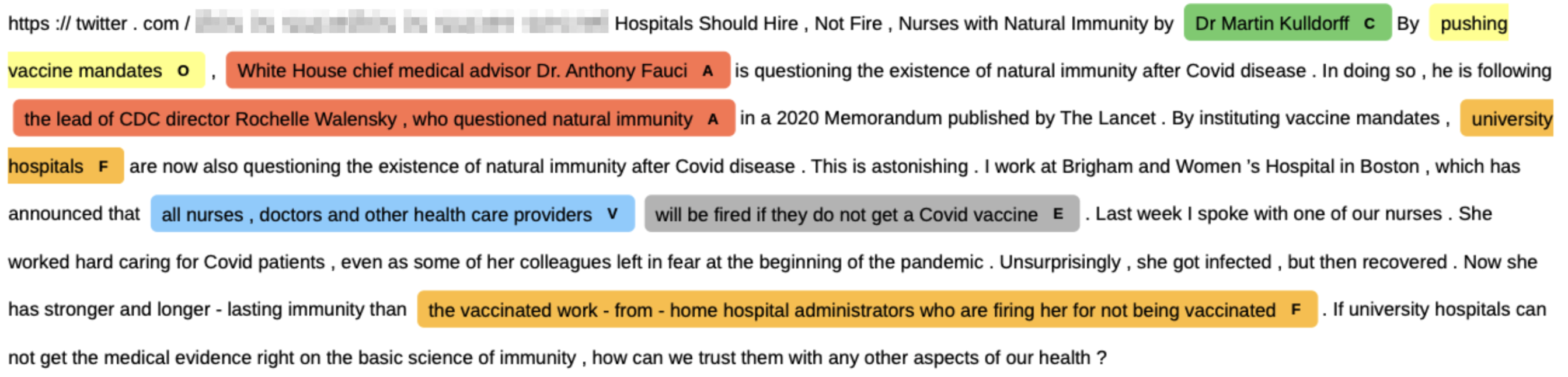} \\
    \end{tabular}
    \caption{A \textit{Critical} message annotated with elements of oppositional narrative: \textit{Agents} (A), \textit{Facilitators} (F), \textit{Campaigners} (C), \textit{Victims} (V), \textit{Objectives} (O), \textit{Negative Effects} (E).}    
    \label{fig:critical}
\end{figure*}


We identified the following six categories of narrative elements (see Figure \ref{fig:conspi} foran example annotation of a \textit{Conspiracy} message, and Figure \ref{fig:critical} for an example annotation of a \textit{Critical} message.):

\begin{enumerate}
\item {\textit{Agents} (A):} Those responsible for the actions and/or negative effects described in the comment. In \textit{Conspiracy}, it could be the hidden power that pulls the strings (in Figure \ref{fig:conspi}, "\textit{Private owned WHO}", "\textit{investors like Bill Gates}", "\textit{pharma companies}" and "\textit{very evil beings}"). In \textit{Critical}, it could be the actors that design the mainstream public health policies (in Figure \ref{fig:critical}, "\textit{White House chief medical advisor Dr. Anthony Fauci}" and "\textit{the lead of CDC director Rochelle Walensky, who questioned natural immunity}").
\item{\textit{Facilitators} (F):}  Those who collaborate with the agents and contribute to the execution of their goals. In \textit{Conspiracy}, they could be governments or institutions which, either intentionally or unwittingly, collaborate with the conspirators and help the conspiracy move forward (in Figure \ref{fig:conspi}, "\textit{the world governments ruled by their puppets}", "\textit{their media}", "\textit{the media}" and "\textit{governments}"). 
In \textit{Critical}, they could be healthcare workers, mass media 
or authority figures who abide by governmental instructions (in Figure \ref{fig:critical}, "\textit{university hospitals}" and "\textit{the vaccinated work - from - home hospital administrators who are firing her for not being vaccinated}").
\item{\textit{Campaigners} (C):} Those who oppose the mainstream narrative. In \textit{Conspiracy}, those who know the truth and expose it to society at large (in Figure \ref{fig:conspi}, "\textit{those awake already}"). In \textit{Critical}, those who oppose the enforcement of laws and/or refuse to follow health-related instructions from the authorities (in Figure \ref{fig:critical}, "\textit{Dr Martin Kulldorff}").
\item{\textit{Victims} (V):}  Those who suffer the consequences of the actions and decisions of the agents and/or the facilitators. In \textit{Conspiracy}, the people who are deceived by those in power, and suffer, become ill, lose their freedom, or die as a result of a hidden plan (in Figure \ref{fig:conspi}, "\textit{people}", "\textit{most people}" and "\textit{regular people}"). 
In \textit{Critical}, the people who receive the negative consequences of the actions and the decisions made by those in power, and also suffer, lose their freedom, become ill, or die as a result of wrong decisions (in Figure \ref{fig:critical}, "\textit{all nurses, doctors and other health care providers}").
\item{\textit{Objectives} (O):} The intentions and purposes that the agents are trying to achieve. In \textit{Conspiracy}, the goals of the conspirators (in Figure \ref{fig:conspi}, "\textit{agenda}" and "\textit{destroying us}"). In \textit{Critical}, the goals of public authorities, pharmaceutical companies, organizations, etc (in Figure \ref{fig:critical}, "\textit{pushing vaccine mandates}").
\item{\textit{Negative Effects} (E):} The negative consequences suffered by the victims as a result of the actions and decisions of those in power and/or their collaborators (in Figure \ref{fig:conspi}, "\textit{the constant fear mongering}" and "\textit{pay a hefty price, often with their health, lives, the loss of their loved ones}"; in Figure \ref{fig:critical}, "\textit{will be fired if they do not get a Covid vaccine}"). 
\end{enumerate}

The decision of where each label should start and end is not as straightforward as it may seem; for this reason, we established the following criteria for span delimitation (in the examples given below, the span is enclosed within square brackets, and the initial of each label appears afterwards within round brackets):
\begin{enumerate}
    \item In general, punctuation marks are excluded from the span. For example, ``[...] \textit{like you and everyone else who isn't getting paid by these satanic billionaires and isn't in on the whole thing?} [\textit{The government}](F)?". However, exclamation and question marks thought to be relevant for the interpretation of the label can be included. For example: [\textit{She leaves behind her year old son… How many more will we lose, how many children?}](E). Also, if an opening punctuation mark has already been included in the span (e.g. quotes, parenthesis, etc.), its corresponding closing punctuation mark must also be included. For example: ``\textit{Why SARS-Cov-2} [\textit{"vaccinated" people}](V) \textit{can test HIV+}". 
    \item Paralinguistic elements, such hashtags, which are considered to be relevant for the interpretation of the comment can be labeled on their own or be included in an adjacent fragment if they both have the same communicative function. For example: "\textit{Another super fit athlete dies suddenly at the age of 35. I wonder what could be causing all these youg, fit athletes to die suddenly? \#COVIDVaccine} [\textit{\#GenocideAgenda}](O)".
    \item The label of categories relative to actors (i.e. \textit{Agents}, \textit{Facilitators}, \textit{Campaigners} and \textit{Victims}) must include the syntactic determiner of the noun phrase (articles, possessive determiners, etc.). For example: ``[\textit{The billionaire Luciferian Elites}](A) \textit{think of themselves as the apex predators of the world and therefore deserving of inheriting the Earth}.".
    \item Each individual element in an enumeration must be annotated separately, with its own label. For example: ``[...] \textit{while} [\textit{Bill Gates}](A), [\textit{the Rockefeller and Ford Foundations}](A), and [\textit{the World Bank}](A) \textit{control 10 per cent of the world’s germplasms} [...]".
    \item Only linguistic elements which have full semantic content can be labeled; thus, purely coreferential elements (personal pronouns, demonstrative pronouns, clitic pronouns, etc.) must not receive their own label unless they are part of a larger phrase which contains semantically full elements. For example: ``[...] \textit{Bill Gates (GAVI, WHO, CDC, Microsoft, most vaccines etc etc) set on destroying and enslaving humanity, by poisoning and starving} [\textit{us, common people}](V)".
    \item In all of these categories referring to an actor of the oppositional narrative, the label must include those elements which help to identify and narrow down the referent (adjectives, defining relative clauses, etc.). For example: ``[...] \textit{the soldiers of} [\textit{the satanic billionaires who like to call themselves Elites}](A).''.
    
\end{enumerate}


The second annotation task was carried out by two teams (one per language) composed of two linguists each. Both teams were supervised by the same senior annotator, with whom they discussed doubts on a weekly basis. In turn, whenever it was necessary, this senior annotator held further meetings with another senior annotator and the psychologist who created the taxonomy to clarify doubts regarding the implementation of the annotation criteria. These criteria were updated in the annotation guidelines through an iterative process (i.e. we started with a basic set of criteria, which were adapted during the initial periods of the task to make sure they could accurately be implemented onto our dataset). In order to assess the quality of this annotation task, the Gamma ($\gamma$) measure \citep{mathet_unified_2015} was used to calculate the IAA. 
Following our preliminary annotation guidelines, we annotated an initial batch of 300 texts, which yielded an average $\gamma$ of 0.43. After revising and discussing the annotation of this first batch, we fine-tuned the guidelines and annotated a second batch of 300 texts with the updated criteria, which yielded an average gamma of 0.57. Finally, a third batch of 300 annotated texts yielded an average gamma of 0.61, which we deemed acceptable since it is close to or above the average agreement of other highly conceptual span-level annotation schemes \citep{prop-fine-grained_2019, weimer_consistency_2022}. 
After that, the rest of the corpus was annotated following the final version of the annotation guidelines.
Following this final version, the IAA was calculated in batches of 200 texts.
Whenever the average gamma result of a batch was below 0.6, the worst quality messages within this batch were discussed and reannotated in order to guarantee the quality of the annotation. After this reannotation, the final average gamma of the task was of 0.72. Given the difficulty of such a subjective and nuanced task, we deem this result as more than acceptable.

The Gold Standard (GS) of this second annotation task
was automatically created applying the following criteria to resolve cases of disagreement: 
\begin{enumerate}
    \item When the annotators agreed on the label but disagreed in the precise limits of a given span (i.e. there was partial overlapping), the GS consisted of the union of both spans. For example, if Annotator 1 labeled as \textit{Facilitators} the span ``\textit{administrators and superiors}'' and Annotator 2 used the same label but selected the span ``\textit{superiors who reportedly never intended to approve them in the first place}'', the GS was created by joining both spans, i.e. ``\textit{administrators and superiors who reportedly never intended to approve them in the first place}''.  
    \item When only one of the annotators labeled a given span, or when there was a disagreement in the labels for the same span, the GS did not incorporate either annotation. 
\end{enumerate}

\begin{table*}[h!]
\centering
{\footnotesize
\begin{tabular}{llcccccc}
\toprule
 &  & A & F & C & V & O & E \\ 
 \midrule
\multirow{3}{*}{ES} & All & 3,329 (14.0\%) & 2,688 (11.3\%) & 4,231 (17.8\%) & 5,260 (22.2\%) & 622 (2.6\%) & 7,150 (30.2\%) \\ 
 & Conspiracy & 1,361 (9.8\%) & 1,184 (8.6\%) & 2,133 (15.4\%) & 3,543 (25.6\%) & 23 (0.2\%) & 5,326 (38.5\%) \\ 
 & Critical & 1,968 (20.0\%) & 1,504 (15.2\%) & 2,098 (21.3\%) & 1,717 (17.4\%) & 599 (6.1\%) & 1,824 (18.5\%) \\ 
\midrule
\multirow{3}{*}{EN} & All & 6,411 (22.4\%) & 3,462 (12.1\%) & 6,416 (22.4\%) & 4,433 (15.5\%) & 2,073 (7.2\%) & 5,565 (19.4\%) \\ 
 & Conspiracy & 3,333 (21.1\%) & 1,336 (8.5\%) & 3,839 (24.4\%) & 2,734 (17.3\%) & 615 (3.9\%) & 3,708 (23.5\%) \\ 
 & Critical & 3,078 (23.9\%) & 2,126 (16.5\%) & 2,577 (20.0\%) & 1,699 (13.2\%) & 1,458 (11.3\%) & 1,857 (14.4\%) \\ 
 \midrule
\end{tabular}}
\caption{Statistics for the gold span-level annotations of the narrative elements. Absolute number and percentage of spans is given, for each of the binary text classes and for all texts, and for each of the six narrative categories: \textit{Agents} (A), \textit{Facilitators} (F), \textit{Campaigners} (C), \textit{Victims} (V), \textit{Objectives} (O), \textit{Negative Effects} (E).}
\label{table:spanstats}
\end{table*}

Table \ref{table:spanstats} shows the amount and the percentages of spans in the GS that have been annotated with each label for each category (\textit{Conspiracy} or \textit{Critical}). 
The low number of certain labels in this table might be due to the way the GS was created, 
especially in spans that can often be interpreted either as \textit{Agent} or \textit{Facilitator} in many messages, which results in label disagreement. The application of criterion (2) results in these instances not being present in the GS.  
Following the learning with disagreement paradigm \citep{Uma2021LearningFD} we provide not only the GS but also the individual annotations of each annotator.

\subsection{Anonymization}
\label{sec:anonymization}
The final step in the creation of the XAI-DisInfodemic corpus was the anonymization of sensitive personal data contained in the messages. Personal data can be proper nouns of anonymous people (i.e. not public figures), phone numbers, email addresses, bank account numbers, and mentions. The procedure we implemented is a combination of automatic and manual approaches:

\begin{enumerate}
    \item Phone numbers and bank account numbers were automatically anonymized. 
    \item Mentions (@username) and emails (username@domain.com) were automatically extracted, and a senior annotator who was familiar with the corpus decided which usernames needed to be anonymized (i.e. those belonging to private individuals, rather than public figures or Telegram channels). The anonymization process consisted in replacing the characters of an username with a randomly generated sequence of alphanumeric characters, with two random numbers added at the end. 
    \item Proper nouns were selected automatically, using the spaCy tool\footnote{\url{https://spacy.io/}}, and reviewed manually by the same senior annotator, taking into account the context in which they appeared. For each proper noun that needed to be anonymized, an alternative replacement was manually given, so that gender and format (e.g. first name, full name, short versions of a name, etc.) were maintained. This step is crucial to ensure that co-reference relationships and gender agreement are kept within each message, so that the cohesion of the discourse is not lost.
\end{enumerate}

\newcommand{\best}[1]{\textbf{#1}}

\section{Predictive Machine Learning}

In this section we describe a series of machine learning experiments aimed at predicting text-related variables provided by the annotation process described in Sections \ref{sec:binaryannot} and \ref{sec:oppositional-annot}. These experiments include the binary classification of texts into the \textit{Critical} and \textit{Conspiracy} categories, the detection of word spans representing the elements of oppositional narratives, and the text-level binary classification of the occurrences of the narrative categories.

Our aim was to assess the performance of standard NLP models and to offer robust solutions for researchers as a basis for future improvements. We view these experiments as a first step towards full automatization of manual annotation in future applications, and we leave the exhaustive exploration of many possible improvements for future work.

\label{sec:ml-experiments}

\subsection{Classifying Critical vs. Conspiracy}
\label{sec:binary}

\newcommand{\hhfone}{human-vs-human}
\newcommand{\hgoldfone}{human-vs-gold}

We first tackled the task of distinguishing between critical and conspiratorial messages, set as a binary classification problem (with \textit{Conspiracy} as the positive class\footnote{We note that, in our experiment, the performance of the transformer-based classifiers is invariant to the definition of positive class: the results (rounded to two decimal places) remain the same when \textit{Critical} is the positive class.}) on Gold Standard data. We experimented with BERT \citep{devlin_bert_2019} and RoBERTa \citep{liu2019roberta} models, using the original English models and their Spanish counterparts, BETO \citep{canete_spanish_2023} and BERTIN \citep{rosa_bertin_2022}. RoBERTa builds upon BERT by using an improved training procedure and more data. 
For English messages, we also experimented with DeBERTaV3 \citep{he_debertav3_2022}, an advanced model that builds upon the BERT/RoBERTa architecture by using improved pre-training and by disentangling the content and the position attention.  

We built the classifiers by fine-tuning the transformer models from the huggingface\footnote{\url{https://huggingface.co/models}} repository, and we used case-sensitive ``base'' versions. The number of tokens was set to 256. We tuned the models for 3 epochs using the AdamW optimizer, learning rate of $2e^{-5}$, slanted triangular LR scheduler with a 10\% warm-up period, a batch size of 16, and a weight decay of 0.01. All the layers of the transformers were fine-tuned. The dropout rate for the classification head was 0.1. In order to obtain robust performance estimates, the results were computed using five-fold cross-validation (4K train and 1K test texts), with separate per-fold random seeds for initialization of model parameters.

 \begin{table}[ht]
 \centering
   \begin{tabular}{lcccc}
     \toprule
     & \multicolumn{2}{c}{English} & \multicolumn{2}{c}{Spanish} \\
     \cmidrule(lr){2-3} \cmidrule(lr){4-5}
     model & F1-CN & F1-CR & F1-CN & F1-CR \\
     \midrule
      BERT & 0.84 & 0.92 & \best{0.80} & \best{0.89} \\   
      RoBERTa & 0.86 & 0.93 & 0.78 & 0.88 \\
      DeBERTa  & \best{0.88} & \best{0.94} &  & \\        
    \midrule
      \hhfone & 0.82 & 0.90 & 0.86 & 0.91 \\
      \hgoldfone & 0.91 & 0.95 & 0.93 & 0.96 \\           
    \bottomrule
 \end{tabular}
 \caption{Performance of classifiers on the \textit{Critical} (CR) vs. \textit{Conspiracy} (CN) task and agreement between human annotators, in terms of the binary F1 measure.}
 \label{table:binarycc}
 \end{table}

In order to compare the performance of models with expected human agreement, we computed the agreement between annotators in terms of the binary F1 score. First, the annotations (Section \ref{sec:binary}) were used to extract binary \textit{Critical} vs. \textit{Conspiracy} labels for each annotator. Then these labels were viewed as class predictions and evaluated against two definitions of ground truth: labels of other annotators (\hhfone), and the labels obtained by the majority vote (\hgoldfone). In each case, the final score was formed by averaging over all possible (prediction, ground truth) pairs. Note that the \hhfone{} scores are pessimistic estimates of human performance, since labels of a single annotator represent a ground truth that is more variable than the average. On the other hand, the \hgoldfone{} scores are optimistic estimates since the individual predictions are merged into the average, and can therefore influence it. 

Table \ref{table:binarycc} contains both the models' and the inter-annotator binary F1 scores. These results show that the classifiers performed well both for Spanish and English, with the English models achieving higher F1 scores. 
Detection of \textit{Critical} messages achieves very high scores for all the models and languages, while the \textit{Conspiracy} scores are somewhat lower, especially for Spanish.
For English, DeBERTaV3 performed best, while for Spanish BERT outperformed RoBERTa. 

When compared to human scores, all the English models outperform the (pessimistic) \hhfone{} scores and stay below the (optimistic) \hgoldfone{} scores. The best-performing DeBERTa model is close to the \hgoldfone{} scores. For Spanish, the models' scores are below both the \hhfone{} and \hgoldfone{} scores, with the performance on critical texts being much closer to the human thresholds than the performance of conspiracy texts. 

Possible reasons for variation in the models' performance across languages are the differences in quality between English and Spanish transformer models, and the differences in language structure, which could make the Spanish classification task more challenging for transformers. The slightly higher human F1 scores for Spanish suggest that the Spanish labeling task is not harder for humans, which is also demonstrated by the IAA scores in Section \ref{sec:binaryannot}.

Confusion matrices of the best-performing models for English (DeBERTa) and Spanish (BERT), displayed in Table \ref{tab:confusion}, offer more insight into performance differences between the \textit{Conspiracy} and \textit{Critical} classes. For both languages, the precision for the \textit{Conspiracy} class is lower than the precision for the \textit{Critical} class. For English, the precision of \textit{Conspiracy} detection is 88\% compared to 95\% for \textit{Critical}  detection, while for Spanish the corresponding precision scores are 78\% and 90\%. In other words, models confuse conspiratorial messages as critical messages more often than they confuse critical messages as conspiratorial, and this asymmetry is more pronounced for Spanish. This is likely due to the fact that conspiratorial messages can be subtle and indirect in their support for conspiracy theories and are therefore more likely to be confused as critical messages. Interestingly, the IAA scores in Section \ref{sec:binaryannot} indicate that human agreement on messages being conspiratorial is higher than on messages being critical. This indicates that humans are able to reliably detect conspiracies based on subtle and indirect cues. Therefore, a reasonable first step in improving the \textit{Conspiracy} vs. \textit{Critical} classifiers would be an analysis of misclassified conspiratorial messages and the identification of the features that are reliable signals of conspiratorial discourse.

\begin{table}[h]
\centering
\begin{tabular}{lcccc}
\toprule
& \multicolumn{2}{c}{English} & \multicolumn{2}{c}{Spanish} \\ 
\cmidrule(lr){2-3} \cmidrule(lr){4-5}
predicted & conspiracy & critical & conspiracy & critical \\ 
\midrule
conspiracy & 301.60 & 35.20 & 284.60 & 60.80 \\ 
critical & 43.20 & 620.00 & 81.00 & 573.60 \\
\bottomrule
\end{tabular}
\caption{Confusion matrices (averaged over the five cross-validation folds) for classification of texts into \textit{Conspiracy} and \textit{Critical} classes, for the best-performing models: DeBERTa for English and BERT for Spanish.}
\label{tab:confusion}
\end{table}

To summarize, the previous results demonstrate that the \textit{Critical} vs. \textit{Conspiracy} task is tractable, and that the expected performance of transformer classifiers is good for Spanish and very good for English, which is an expected result of combining high-quality annotations with strong transformer classifiers. High F1 scores for the detection of critical messages suggest that these classifiers could be reliably applied to make the automatic conspiracy-filtering tools more fair. In this use-case, the texts labeled as \textit{Critical} could either be automatically unflagged as \textit{Conspiracy}, or routed to humans for closer inspection.

\subsection{Detection of Narrative Elements}
\label{sec:seqlabel}

We proceeded with the identification, on the fine-grained level of token spans, of the six elements of  oppositional narratives defined in our annotation scheme (for details, see Section \ref{sec:binaryannot}). Our task includes both long and overlapping spans (of different categories). Similar tasks have been carried out in the context of propaganda detection \citep{prop-fine-grained_2019, martino_semeval2020_toxicity}, skill extraction \citep{zhang2022skillspan}, and the analysis of literary texts \citep{weimer_literaryspan_2022}.

For evaluation, we used the adapted F1 measure of \cite{prop-fine-grained_2019} which accounts for partially correct predictions by looking at span overlap (span-F1). This approach offers a fairer evaluation in tasks with long spans, and with inherent subjectivity of the span boundaries. For tasks like traditional non-nested NER, where named entities are shorter and expectedly have well-defined boundaries, exact matching is a reasonable method of evaluation.

We approached the task by fine-tuning transformer models with token classification heads. To account for the possibility of overlapping spans with different categories, we used 6 separate per-category heads that performed BIO sequence tagging. We employed multi-task learning \citep{ruder_overview_2017} by connecting the per-category taggers to the same transformer backbone. Multi-task learning has several advantages such as improved regularization and implicit data augmentation \citep{ruder_overview_2017}, and the described approach was successfully deployed for a similar task of span-level skill extraction \citep{zhang2022skillspan}. 

We experimented with the same transformers used for binary classification (Section \ref{sec:binary}) and we used the same configuration and hyperparameters. The exception was the number of epochs, which we increased to 10 in order to accommodate for the increased task complexity. Gold Standard data was used and it was adapted for multi-task learning by creating a separate training instance for each text and each span category, which resulted in a large proportion of texts without spans. In order to avoid overfitting to the unlabeled tokens, empty texts were downsampled (per category) to 10\% of the original size. 
Five-fold cross-validation was used to obtain performance estimates, in the same way as in Section \ref{sec:binary}.

The results, presented in the top half of Table \ref{table:narrativecat}, show the average and per-category span-F1 scores for both languages. For English, DeBERTaV3 outperforms RoBERTa, which outperforms BERT, while for Spanish RoBERTa outperforms BERT. The per-category performance correlates well with the number of spans annotated with a category, i.e. with the size of the training data for the corresponding token classifier. For example, the (less frequent) \textit{Facilitators} and \textit{Objectives} categories are harder to detect. Additionally, the Spanish BERT and RoBERTa models are better than their English counterparts for the \textit{Victims} and \textit{Negative Effects} categories which are more frequent in Spanish, while for most of the other categories the English models work better. For English, the most sophisticated DeBERTaV3 model achieves noticeable performance gains for most of the categories.

Additionally, we experimented with the SpanBERT model \citep{joshi_sbert_2020} as the transformer backbone. SpanBERT is a model based on a pre-training method  ``designed to better represent and predict spans of text'' \citep{joshi_sbert_2020}. The SpanBERT model achieved an inferior performance of 0.31 average span-F1 on English texts (Spanish SpanBERT model does not exist). This is in line with the skill extraction experiment \citep{zhang2022skillspan}, in which the SpanBERT model was not the best solution and it rarely outperformed BERT.

Finally, we experimented using CRF \citep{lafferty-2001-crf}, a probabilistic model that learns patterns of transition between token classes\footnote{We used the \texttt{pytorch-crf} package: \url{https://pytorch-crf.readthedocs.io}}, as the final token classification layer. We experimented with two CRF-based approaches: one in which the CRF layers were optimized together with the transformer (end-to-end), and the other in which the CRF layers were optimized after training and freezing the transformer (fine-tuned). Both of the CRF-based approaches led to an inferior performance. In terms of the span-F1 measure, the average (end-to-end CRF, fine-tuned CRF) results  for English were: (0.23, 0.19) for BERT, (0.37, 0.36) for RoBERTa, and  (0.47, 0.46) for DeBERTa. For Spanish, the results were (0.32, 0.38) for BERT, and  (0.23, 0.25) for RoBERTa. The likely reason for the inferior performance of CRF is that its inductive bias (an assumption that there is valuable information in the surface patterns of transition between token tags) does not fit well with the nature of the data in our case.
This is supported by the average number of spans of the same category per text (in which the category occurs), which is 2.29 for English and 2.39 for Spanish.

In terms of the results on a different but similar task, best span labeling models detecting a single ``propaganda'' category in English news articles achieved a span-F1 of 0.52 \citep{martino_semeval2020_toxicity}. 
While direct comparison is not possible, we take these results as evidence that our approach (average span-F1 of 0.57 for English and 0.50 for Spanish) achieves solid performances on a challenging task. 

Regarding \rqtwo, we can say that while the performance of the proposed approach is not below the top performance expected from a similar task, there is room for further improvement, especially for the less frequent categories. Techniques that could improve performance on the less frequent categories include data augmentation \citep{feng-etal-2021-survey} and active learning \citep{erdmann-etal-2019-practical}. 

\begin{sidewaystable}
 \centering
 {
   \begin{tabular}{clcccccccccccccc}
    \toprule
     & & \multicolumn{7}{c}{English} & \multicolumn{7}{c}{Spanish} \\
     \cmidrule(lr){3-9} \cmidrule(lr){10-16}
     & & Avg. & A & F & C & V & O & E & Avg. & A & F & C & V & O & E \\   
     \midrule
     & num. spans & 4,727 & 6,411 & 3,462 & 6,416 & 4,433 & 2,073 & 5,565 & 3,880 & 3,329 & 2,688 & 4,231 & 5,260 & 622 & 7,150 \\          
     \midrule     
    \multirow{3}{*}{\rotatebox[origin=c]{90}{SPAN}} 
      & BERT & 0.50 & 0.59 & 0.38 & 0.54 & 0.57 & 0.38 & 0.54 & 0.48 & 0.46 & \best{0.39} & 0.50 & 0.57 & 0.33 & 0.64 \\         
      & RoBERTa & 0.52 & 0.60 & 0.41 & \best{0.62} & 0.58 & 0.36 & 0.55 & \best{0.50} & \best{0.51} & 0.37 & \best{0.54} & \best{0.62} & \best{0.33} & \best{0.64} \\
      & DeBERTa & \best{0.57} & \best{0.63} & \best{0.45} & 0.61 & \best{0.63} & \best{0.49} & \best{0.62} &  & & & & & & \\       
    \midrule    
    \multirow{3}{*}{\rotatebox[origin=c]{90}{BINARY}}
      & BERT & 0.73 & 0.82 & 0.60 & 0.79 & 0.77 & 0.64 & 0.78 & \best{0.72} & \best{0.72} & \best{0.60} & \best{0.75} & \best{0.83} & \best{0.59} & \best{0.85} \\         
      & RoBERTa & 0.76 & 0.84 & 0.64 & 0.82 & 0.80 & 0.66 & 0.80 & 0.69 & 0.71 & 0.58 & 0.74 & 0.83 & 0.57 & 0.84 \\
      & DeBERTa & \best{0.77} & \best{0.84} & \best{0.65} & \best{0.83} & \best{0.80} & \best{0.69} & \best{0.81} & & & & & & &  \\ 
    \bottomrule 
 \end{tabular}}
 \caption{Performance of BERT, RoBERTa, and DeBERTaV3 transformers on the tasks of span labeling (SPAN), and binary detection (BINARY) of the text-level narrative categories, in terms of span-F1 and binary F1 measures, respectively. Total number of spans in the dataset for each of the categories is given in the top row. The categories are: \textit{Agents} (A), \textit{Facilitators} (F), \textit{Campaigners} (C), \textit{Victims} (V), \textit{Objectives} (O), \textit{Negative Effects} (E).}
 \label{table:narrativecat}
\end{sidewaystable}

\subsection{Detection of Narrative Elements}

While narrative categories in Section \ref{sec:seqlabel} were detected on the token level, here we will tackle their text-level detection. For each category we assigned binary labels to texts, labeling a text as positive if the category was present in the text, i.e.  represented by at least one Gold Standard span. Such per-category binary variables can convey useful information: the presence of spans labeled as \textit{Facilitator} or \textit{Campaigner} is a signal of inter-group conflict (for details, see Section \ref{sec:css-experiments}).

The models and their configuration are the same as for the \textit{Conspiracy} vs. \textit{Critical} (CC) classification in Section \ref{sec:binary}, as well as the training and evaluation. The results in Table \ref{table:narrativecat} show that binary detection of the narrative categories is, on average and for most of the categories, a harder task than CC classification (Table \ref{table:binarycc}). Relative model performances are the same: English models obtain better results than Spanish, with DeBERTaV3 being the most successful, and BERT outperforms RoBERTa in Spanish. 

The results in Table \ref{table:narrativecat} also show that the binary per-category scores correlate very well with the per-category scores of the sequence labeling models.
This is expected since the per-category labeling scores correlate well with the number of spans annotated with a category, which correlates with the number of positive examples for the per-category classification. Similarly as for sequence labeling, the  smallest \textit{Facilitators} and \textit{Objectives} categories are the most challenging ones and have similar scores.  

Regarding \rqthree, these results show that text-level detection is a more viable option in applications that require this level or granularity, and that good results can be achieved for most narrative elements. However, the \textit{Facilitators} and \textit{Objectives} categories remain a challenge. One could likely improve performance on these less frequent categories by using techniques such as data augmentation \cite{feng-etal-2021-survey} and active learning \cite{settles-2011-closing}.

\subsection{Error Analysis}

The results indicate that the \textit{Facilitators} (F) and \textit{Objectives} (O) categories are the most challenging to detect (Table \ref{table:narrativecat}). We focus the error analysis on the \textit{Facilitators} span since it conveys important information about inter-group conflict (Sections \ref{sec:dataset}, \ref{sec:css-experiments}). 
We hypothesize that the model confuses the \textit{Agent} (A) and \textit{Facilitator} (F) labels, which have a similar role in terms of the ``friend vs. foe'' discourse. To examine their relationship, we introduce, for each of the categories, a text-level variable that codifies classification outcomes: \textit{False Negative}, \textit{Correct} and \textit{False Positive}. For both categories and for each language, we calculated the outcomes of the best-performing classifiers (see Table \ref{tab:erroranal}). Predictions for all five folds are pooled to cover the entire dataset. Applying the chi-square test, which assesses the independence between two categorical variables, we found that the classification outcomes for \textit{Facilitators} (F) are not independent of the outcomes for \textit{Agent} (A) ($p < 0.001$). 

Next we perform the analysis of the residuals, which measure how much a particular combination of outcomes influences the dependence of the variables -- large residuals influence the dependence more, and vice versa \citep{Sharpe2015}. Table \ref{tab:erroranal} shows that there are 81 English texts with missed \textit{Facilitators} and falsely detected \textit{Agents}, and that this combination strongly influences the dependence between variables (high residual of 4.3).
In our context, the values of the residuals show that there is a strong correlation between \textit{Facilitators} being missed (False Negative) and \textit{Agents} being incorrectly identified (False Positive), for both languages. For Spanish, incorrectly identifying \textit{Facilitators} (False Positive) was significantly linked to missing \textit{Agents} (False Negative). These results show a tendency to misclassify \textit{Facilitators} as \textit{Agents} in both languages, and to misclassify \textit{Agents} as \textit{Facilitators} in Spanish. Future models should tackle this confusion, possibly by applying a multi-task approach with a loss function modified to penalize the problematic cases. 

\begin{table}[h!]
    \centering
    \begin{tabular}{lrrr} 
    \hline
    English & A-FN & A-Corr & A-FP \\     
    \hline
    F-FN & 24 (-2.0) & 418 (-2.1) & 81 (4.3) \\ 
    F-Corr & 294 (2.6) & 3,452 (1.6) & 384 (-4.1) \\ 
    F-FP & 17 (-1.4) & 289 (0.2) & 40 (0.9) \\ 
    \hline
    Spanish & A-FN & A-Corr & A-FP \\ 
    \hline
    F-FN & 46 (0.7) & 481 (-3.1) & 65 (3.5) \\ 
    F-Corr & 268 (-2.6) & 3,514 (5.3) & 269 (-4.6) \\ 
    F-FP & 40 (3.2) & 278 (-4.2) & 39 (2.6) \\ 
    \hline
    \end{tabular}
    \caption{Counts and adjusted residuals (in brackets) for classification outcome variables in the detection of \textit{Facilitators} (F) and \textit{Agents} (A). The outcomes are False Negatives (FN), Correct Classification (Corr), and False Positives (FP).}
    \label{tab:erroranal}
\end{table}

\section{Applying the Annotation Scheme}
\label{sec:css-experiments}

\newcommand{\hypoA}{\textbf{[H1]}\xspace}
\newcommand{\hypoB}{\textbf{[H2]}\xspace}
\newcommand{\hypoC}{\textbf{[H3]}\xspace}
\newcommand{\hypoD}{\textbf{[H4]}\xspace}

In this section, we illustrate how the proposed annotation scheme can be used for interdisciplinary research in computational linguistics and computational social sciences. First we do a proof-of-concept analysis of the taxonomy to demonstrate that conspiratorial and critical discourses differ in their potential effects on reality by measuring the presence of anger and violence in written texts. 
Anger is a core emotion that prepares individuals to correct perceived offenses and take action in violent riots \citep{Zeitzoff_2013}. We hypothesize that conspiracy messages differ from critical messages because they tend to create a specific emotional atmosphere with the use of anger \hypoA and words connected with violence \hypoB.
 
Secondly, we focus on understanding the role of IGC as a powerful force driving violent mobilizations \citep{Moskalenko_08}. This experiment is motivated by the suggestion that CTs are not just pieces of disinformation but narrative artifacts aimed to reinforce group dynamics and coordinate actions \citep{WAGNEREGGER2022}. We hypothesize that a greater degree of IGC will be found in conspiratorial messages \hypoC, and that the use of angry and violent words will be higher in messages where the IGC is present \hypoD. 

To test the previous hypotheses, we define and compute three new variables: \textit{anger}, \textit{violence} and \textit{intergroup conflict}. 

\textbf{Anger:} We calculate the amount of angry words in a messasge by applying the LIWC2007 Spanish dictionary \citep{ramirez_liwc_es_2007} and the LIWC2015 English dictionary \citep{pennebaker_liwc_en_2015}. The LIWC dictionary is a standard method in psychological and linguistic research that groups words in emotional and cognitive categories. We define the \textit{anger} variable as the percentage of words in a text that belong to the \textit{anger} category in LIWC.

\textbf{Violence:} We aim to measure discourse that promotes political violence (PV) \citep{Bosi2015, kruger2018fomenting}. Based on \citet{kruger2018fomenting}, we define this kind of discourse as the pragmatic use of a specific vocabulary related to aggression that implicitly normalizes or legitimizes violent acts in order to achieve specific goals. Therefore, we opt for a dictionary-based approach similar to the one used for anger.

The challenge of this approach is to obtain both English and Spanish words that relate specifically to political violence and do so in the context of COVID-19 discourse (e.g. words related to death and injury can merely discuss the pandemic). Therefore, we use a ChatGPT-based method to obtain a list of context-specific PV words for both languages. The method consists of two steps: 1) creation and application of a language-specific ChatGPT prompt; 2) curation of the list by human experts. The prompt asks ChatGPT to extract, from a single message, a list of words that create a violent and aggressive atmosphere. For each language, we applied the corresponding prompt to a sample of 1,000 messages. The output words were lemmatized and checked by two experts, which resulted in a total of 587 Spanish and 710 English words. The words were then used to calculate the percentages of violent words in messages. We have manually examined a sample of messages in order to validate the indices of anger or violence. The prompts, the list of words, and examples of violent messages are provided as supplementary material. The correlation between the presence of words from the two lexicons in the messages of the whole corpus is significant but moderate (R=.49 for English and R=.37 for Spanish). This suggests, as we expected, that the promotion of violence and the expression of anger are related but not entirely equivalent phenomena.

\textbf{Intergroup conflict:} In order to describe the presence of a narrative that focuses on Intergroup Conflict, we rely on the presence of \textit{Campaigners} and \textit{Facilitators} in a message (see Section \ref{sec:oppositional-annot} for definitions). These two types of actors are related to IGC because they represent the social groups that are in confrontation with each other in real life. Based on the presence of these categories, we created a categorical variable that identifies four manifestations of IGC: no IGC actors are present in the text (C1), only \textit{Campaigners} are present (C2), only \textit{Facilitators} are present (C3), both \textit{Campaigners} and \textit{Facilitators} are present (C4). While C1 indicates the absence of IGC, C4 indicates a strong presence of IGC, since both groups in conflict are explicitly mentioned.

\subsection{Analysis and Results}

Firstly, we investigate the difference in the amount of \textit{anger} and \textit{violence} words between conspiratorial and critical texts. Since these variables are not normally distributed according to the Kolmogorov-Smirnov test ($p<0.001$), we used the Mann-Whitney U test for comparison of the two groups.
 
\begin{figure}[h]
\includegraphics[width=1.0\textwidth]{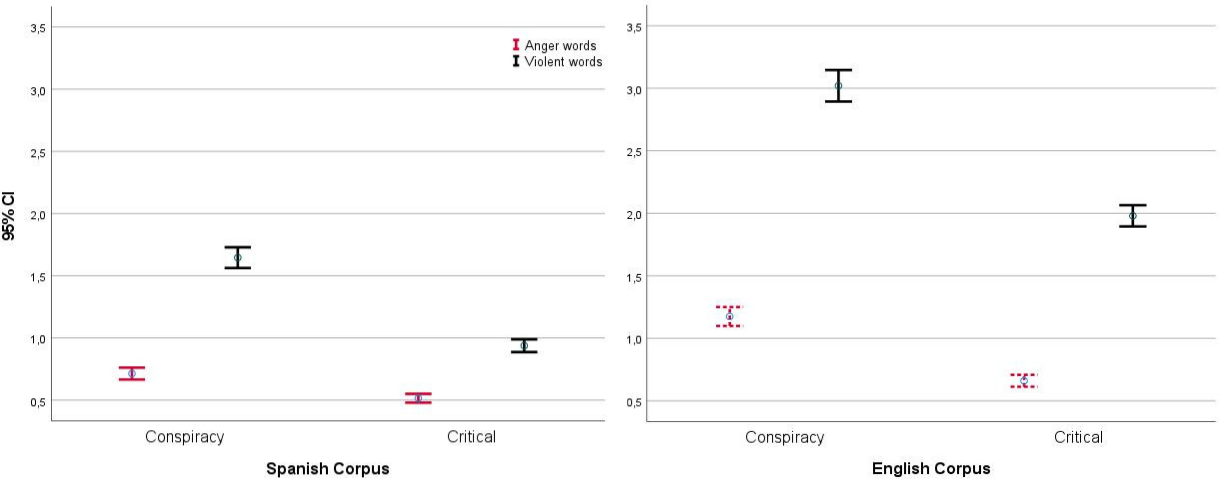}
\caption{Relation of the conspiracy and critical label with the presence of \textit{anger} and \textit{violence} words in texts.}
\label{figure:angerviolencepercateg}
\end{figure}

As we observe in Figure \ref{figure:angerviolencepercateg}, conspiratorial messages convey significantly more anger 
than critical messages, both in Spanish (U=2352490.5, df=3, $p<0.001$) and in English
(U=2046581.5, df=3, $p<0.001$), which confirms \hypoA. 
Similarly, we observe that conspiratorial messages convey significantly more violent words 
both in Spanish (U=2046572.5, df=3, $p<0.001$) and in English (U=2030201.5, df=3,
$p<0.001$), which confirms \hypoB. 

Next we test whether a larger degree of IGC is present in conspiratorial messages than in critical messages. To this end, we perform a chi-square test and a residual analysis which helps to interpret the chi-square test: the size of the residual indicates a major contribution of a cell to the chi-square significance (for more details, see \citet{Sharpe2015}). 
The results confirm a significant relation between the IGC variable, and the \textit{conspiracy} vs. \textit{critical} variable, both for Spanish ($\chi^2$ = 182.02, df = 3, $p<0.001$) and for English  ($\chi^2$ = 159.16, df = 3, 
$p<0.001$). 
For both languages, the C4 category (which is a clear indicator of IGC) occurs significantly more in conspiracy than in critical messages, while the opposite is true for C1 (indicator of no IGC), which supports \hypoC. Table \ref{table:igc-cc-merged} contains the statistics for the combined texts of both languages.

\begin{table}[ht]
\centering
\begin{tabular}{lccc}
\hline
          & Critical               & Conspiracy           & All \\
\hline
C1       & 43.91\% (9.6)         & 34.11\% (-9.6)    & 40.44\% \\
C2       & 34.34\% (3.8)         & 30.63\% (-3.8)    & 33.02\% \\
C3       & 12.27\% (-7.2)          & 17.47\% (7.2)   & 14.12\% \\
C4       & 9.46\% (-12.1)          & 17.77\% (12.1)  & 12.41\% \\
\hline
\end{tabular}
\caption{Percentages (and residuals) of IGC categories: in critical, conspiracy, and all texts.}
\label{table:igc-cc-merged}
\end{table}

\begin{figure}[h]
\centering
\includegraphics[width=0.47\textwidth]{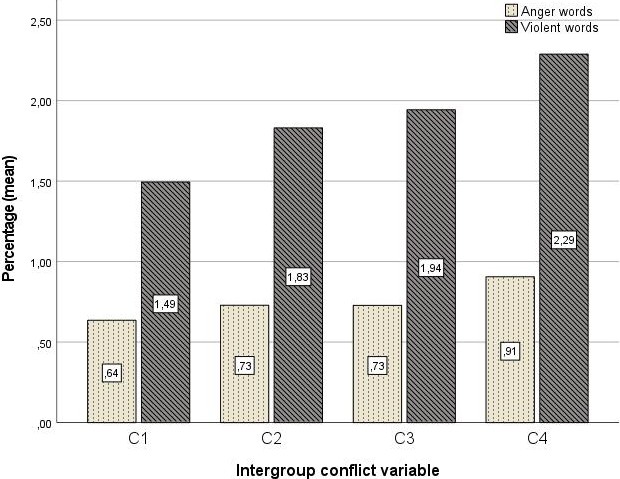}
\caption{Relation of the IGC variable with the percentage of \textit{anger} and \textit{violence} words in messages.}
\label{figure:css1}
\end{figure}

Finally, we tested whether the use of \textit{anger} and \textit{violence} words is higher in messages in which IGC is expressed. We use the Kruskal-Wallis test to compare the presence of \textit{anger} and \textit{violence} words in the different categories of the IGC variable. We observed a statistically significant effect of the IGC variable on the measures of \textit{anger} and \textit{violence} words, both in Spanish (H=96.289, df=3, 
$p<0.001$), and in English (H=79.462, df=3, $p<0.001$). 
As we can see in Figure \ref{figure:css1}, there is a significantly larger amount of both \textit{anger} and \textit{violence} words in messages which have the highest IGC (C4), and a lesser amount 
in the messages which have the lowest IGC (C1), supporting \hypoD. 

\section{Conclusions and Future Work}
\label{sec:conclusions}
This research introduces a new topic-agnostic annotation scheme to approach the analysis of oppositional narratives (\textit{conspiracy} vs. \textit{critical}) and the presence of intergroup conflict in these narratives. With this scheme, we created the XAI-DisInfodemic corpus, consisting of novel English and Spanish datasets of Telegram messages related to the COVID-19 pandemic. To the best of our knowledge, this is the first corpus addressing the important issue of differentiating between conspiratorial and critical messages, and may lead to the improvement of content moderating systems and to the mitigation of censorship. 

The machine learning experiments we conducted show that critical and conspiratorial messages can be differentiated with a high level of accuracy. Span-level detection of oppositional narrative elements has proven to be a more challenging task. We provide strong baseline solutions that should be improved in future work. Possible methods of improvement include advanced multi-task learning methods \citep{ruder_overview_2017}, the use of multi-lingual models, and the creation of transformers adapted to Telegram messages. The results of the text analysis experiments, based on the annotation scheme, demonstrate that conspiratorial discourse is more harmful since it contains higher levels of anger and political violence. These results support the conclusion that censoring critical messages as if they were conspiratorial is to ignore that these messages represent a discourse of a different nature, since messages containing conspiracy theories are more likely to promote emotions such as anger and to use words directly related to violence than critical messages. A future line of research in the area of social psychology should investigate whether these different language features translate into different effects on the audience.

A potential application of these findings is the use of measures for identifying anger and violence in messages in order to improve the decisions of the classifiers and their explainability. Our experiments could also motivate further research on the dissemination of conspiracy theories, since it would be both theoretically and practically relevant to determine if the presence of IGC is related to an increase in the possibility that a given text becomes viral. 

An important direction for future research is to examine whether NLP models can apply our annotation scheme with an accuracy that is high enough to support the findings derived from human-annotated messages. Our experiments and analyses are a first step in this direction.

\section*{Ethics Statement}
This work focuses on differentiating conspiratorial messages from critical messages, which may contribute to the mitigation of unfair censorship. Furthermore, we develop the understanding of oppositional narratives by proposing technical and conceptual tools for the analysis of conspiracy theories, critical texts, and intergroup conflict online. Therefore, our work has the potential to lead to a better understanding of important social challenges, to mitigate adverse effects of conspiracies and censorship, and to contribute to building a safer online environment.

The messages used in our work have been anonymized removing personal data such as names, telephone numbers, addresses, and bank accounts.

Transformer models are prone to biases that can lead to an unfair or harmful misclassification of critical messages as conspiratorial, and vice versa. Potential researchers should therefore take steps to mitigate such biases by performing the appropriate evaluations and model corrections. A related issue is the definition of \textit{Critical} and \textit{Conspiracy} classes, which should be carefully distinguished and correctly identified in order to avoid potential censorship.

Training transformers from scratch consumes a considerable amount of energy and contributes to global warming. The approaches that we propose rely on fine-tuning, which is less of an issue since it consumes much less energy. However, if the implementation of such approaches to new text domains or languages requires the construction of new transformers, we encourage researchers and professionals from private sectors to take into account the environmental consequences.

\section*{Limitations}

Even though we have designed our annotation scheme to be topic agnostic, we have limited our study to the COVID-19 pandemic. Therefore, in order to  validate it even further, it could be applied to other controversial topics that are prone to CTs, such as climate change.

We have focused on messages posted on the Telegram platform, and it remains to be tested to what extent our approach is generalizable to texts from other sources. We have only tested our approach on messages written in English and Spanish, two languages which are highly-resourced and belong to a similar cultural framework. Further experiments are needed to assess the universality of our annotation scheme.

Another limitation of our work is that we have not performed a hyperparameter tuning, but we have chosen to rely on hyperparameters that are commonly used to fine-tune a pre-trained model.

Finally, while we are aware of the importance of taking into account the different perspectives of the annotators, we have not used the learning with disagreement paradigm \citep{Uma2021LearningFD} to leverage them in our experiments. We plan to do this as part of future research.

\section*{Data and Code Availability}
We make the code of the experiments, as well as the data, available.\footnote{\repo} 

\section*{Acknowledgements}

This work was done in the framework of the research project XAI-DisInfodemics:
eXplainable AI for disinformation and conspiracy detection during infodemics
(PLEC2021-007681) funded by MICIU/AEI/ 10.13039/501100011033 and by ``European Union NextGenerationEU/PRTR''.

We would like to thank Mariona Coll Ardanuy for all the helpful advice pertaining to the improvement of the article text and concept.

\section*{Conflict of Interest Statement}

The authors declare no potential conflict of interest.


\bibliographystyle{plainnat}

\bibliography{0-PAPERBIB}

\end{document}